\documentclass[10pt,twocolumn,letterpaper]{article}

\usepackage[pagenumbers]{cvpr} 
\usepackage{multirow}
\usepackage{comment}

\DeclareMathOperator*{\argmin}{arg\,min}

\usepackage[accsupp]{axessibility}  
\usepackage{mathtools}
\DeclarePairedDelimiter\ceil{\lceil}{\rceil}

\definecolor{cvprblue}{rgb}{0.21,0.49,0.74}
\usepackage[pagebackref,breaklinks,colorlinks,allcolors=cvprblue]{hyperref}

\def\paperID{11922} 
\def\confName{CVPR}
\def\confYear{2025}

\title{PAVE: \underline{P}atching and \underline{A}dapting \underline{V}id\underline{e}o Large Language Models}

\author{Zhuoming Liu\textsuperscript{1}, Yiquan Li\textsuperscript{1}, Khoi Duc Nguyen\textsuperscript{1}, Yiwu Zhong\textsuperscript{2}, Yin Li\textsuperscript{1}\\
{\textsuperscript{1}University of Wisconsin-Madison \ \ 
\textsuperscript{2}The Chinese University of Hong Kong}
}

\begin{document}
\maketitle
\begin{abstract}

Pre-trained video large language models (Video LLMs) exhibit remarkable reasoning capabilities, yet adapting these models to new tasks involving additional modalities or data types (e.g., audio or 3D information) remains challenging. In this paper, we present PAVE, a flexible framework for adapting pre-trained Video LLMs to downstream tasks with side-channel signals, such as audio, 3D cues, or multi-view videos.
PAVE introduces lightweight adapters, referred to as ``patches,'' which add a small number of parameters and operations to a base model without modifying its architecture or pre-trained weights. In doing so, PAVE can effectively adapt the pre-trained base model to support diverse downstream tasks, including audio-visual question answering, 3D reasoning, multi-view video recognition, and high frame rate video understanding. Across these tasks, PAVE significantly enhances the performance of the base model, surpassing state-of-the-art task-specific models while incurring a minor cost of $\sim$0.1\% additional FLOPs and parameters. Further, PAVE supports multi-task learning and generalizes well across different Video LLMs. Our code is available at \url{https://github.com/dragonlzm/PAVE}.


\end{abstract}    
\section{Introduction}
\label{sec:intro}


Large multimodal models have recently demonstrated remarkable success in video understanding~\cite{li2024llava, Qwen2VL, gpt4}. These models, known as video large language models (Video LLMs), are pre-trained on video and text data, exhibiting significant reasoning capabilities across video understanding tasks. They thus promise to serve as video foundation models, characterized by their potential to adapt to many downstream tasks~\cite{bommasani2021opportunities}. However, many of these tasks involve \textit{side-channel signals in the form of additional modalities or data types} beyond the video-text pairs that Video LLMs are typically trained on. For example, audio-visual question answering (QA)~\cite{alamri2019audiovisualsceneawaredialog} necessitates integrating both video and audio to comprehend content, while 3D QA~\cite{azuma_2022_CVPR} requires reasoning about the 3D scene from a video of a 3D scan. 
\textit{What if a target video task offers side-channel signals different from the pre-training of Video LLMs?}

\begin{figure}[t!]
    \centering
    \includegraphics[width=0.95\linewidth]{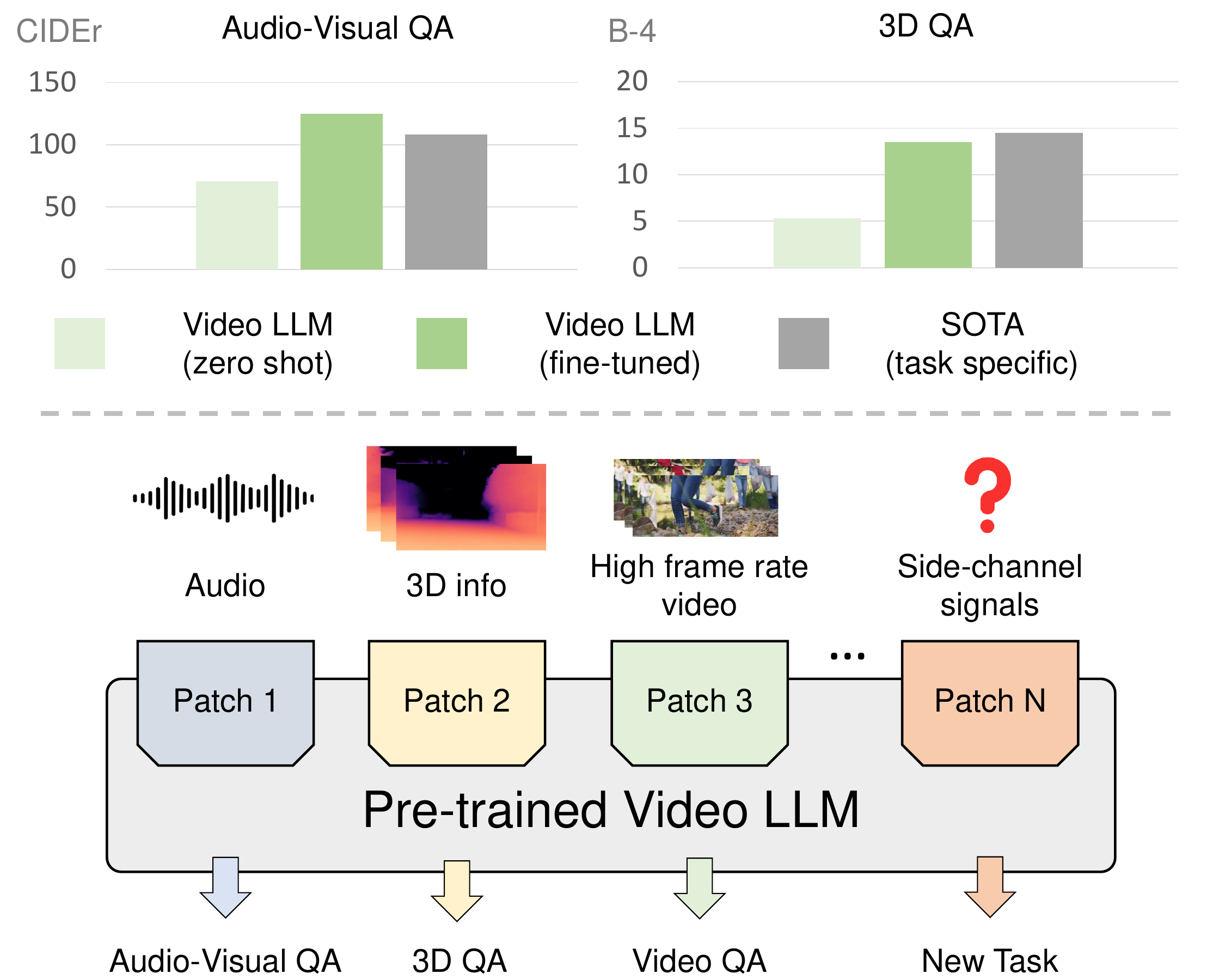}
    \vspace{-0.5em}
    \caption{\textbf{Top}: Evaluating a Video LLM on audio-visual QA and 3D QA tasks. \textbf{Bottom}: Adapting Video LLMs by adding a small ``patch'' of additional operations and parameters, without changing its existing architecture or vast pre-trained weights.}
    \label{fig:teaser}
    \vspace{-1.5em}
\end{figure}

To answer this question, we begin by evaluating the zero-shot performance of a recent video LLM (LLaVA-OneVision~\cite{li2024llava}) on audio-visual QA~\cite{alamri2019audiovisualsceneawaredialog} and 3D QA tasks~\cite{azuma_2022_CVPR}. Our results, as in Figure~\ref{fig:teaser} (top), show that despite lacking access to audio or 3D-specific data, the Video LLM attains promising performance. Yet, with a similar number of parameters, the Video LLM lags behind those task-specific models that leverage additional side-channel signals. Noting that these task-specific models are indeed trained using the dedicated training sets, for a fair comparison, we further fine-tune the Video LLM on the same training sets yet only with video data (see Figure~\ref{fig:teaser} (top)). Surprisingly, this fine-tuned Video LLM approaches or even surpasses the performance of those specialized models with merely video input. 
Motivated by this observation, our key research question is, \textit{can we leverage the knowledge in pre-trained Video LLMs for such tasks?}

To address this question, we investigate the adaptation of pre-trained Video LLMs to downstream tasks with \textit{side-channel signals}, defined as supplementary signals from additional modalities or data types such as audio, 3D cues, multi-view videos, or high frame rate inputs. 
By considering different signals, our formulation addresses key challenges in video understanding. 
For example, considering high frame rate videos as the side-channel explores fundamental questions about video representation~\cite{feichtenhofer2019slowfastnetworksvideorecognition}. Similarly, incorporating multi-view videos, audio, and 3D cues accounts for cross-view and cross-modality reasoning~\cite{grauman2024ego, alamri2019audiovisualsceneawaredialog}.




We propose to adapt a pre-trained base Video LLM through \textit{patching} --- adding a lightweight adapter (``patch'') with a small number of additional parameters and operations, and without altering the base model's architecture or vast pre-trained weights (see Figure~\ref{fig:teaser} (bottom)). Inspired by the success of parameter-efficient adapters (\eg, LoRA~\cite{hu2021loralowrankadaptationlarge}) in text and image generation models (\eg, LLMs~\cite{llama3_2, vicuna2023} and diffusion models~\cite{rombach2022highresolutionimagesynthesislatent, polyak2024moviegencastmedia}), this approach allows for flexible customization, and facilitates convenient sharing of the customization by distributing small patches.

To this end, we present \textbf{PAVE}, a framework designed to \underline{P}atch and \underline{A}dapt \underline{V}id\underline{e}o LLMs with side-channel signals. PAVE leverages cross-attention that operates between tokens derived from key video frames (as queries) and tokens from side-channel signals (as keys and values). This operation aligns the visual signal and side-channel signals along the time axis, fuses the signals from both sources, and then updates the input visual tokens to the LLM. In doing so, PAVE integrates side-channel signals with lightweight ``patches,'' while enabling effective adaptation to various downstream tasks without altering pre-trained models.


To evaluate PAVE, We conduct extensive experiments on four video tasks: (1) audio-visual QA, (2) 3D QA, (3) high frame rate video understanding, and (4) multi-view video recognition. Across all tasks, PAVE effectively adapts a base Video LLM~\cite{li2024llava} and consistently outperforms task-specific models. For example, compared to latest task-specific models, PAVE achieves a relative boost of 2\% on the AVQA~\cite{yang2022avqa} dataset for audio-visual QA, 6\% on the SQA3D dataset~\cite{ma2022sqa3d} for 3D QA, 1-5\% across video benchmarks~\cite{fu2024video, li2023mvbench, zhou2025mlvubenchmarkingmultitasklong} when integrating high frame rate videos, and 1\% on the Ego-Exo4D dataset~\cite{grauman2024ego} for multi-view recognition. In all cases, PAVE only adds about 0.1\% of FLOPs and parameters in addition to the base model. Further, we demonstrate that PAVE can support multi-task learning and generalizes well across different Video LLMs.


\medskip
Our main \textbf{contributions} are summarized as follows.
\begin{itemize}
    \item We present PAVE, a novel framework to adapt pre-trained Video LLMs for tasks with side-channel signals, greatly extending the capacity of Video LLMs.
    \item We design lightweight ``patches'' that add a small number of parameters and operations to a base model, and keep the original architecture and pre-trained weights unchanged, enabling PAVE's adaptation to varying tasks. 
    \item We demonstrate that PAVE can be applied across video tasks and Video LLMs, surpassing the performance of strong task-specific models. 
\end{itemize}

\section{Related Works}
\label{sec:related_works}

\noindent \textbf{Video large language models}.
Recent advances in instruction tuning with visual and text data~\cite{liu2023visualinstructiontuning, liu2023improvedllava, liu2024llavanext} have led to a surge of interest in developing Video LLMs. Many of these models share a common design, where visual features are extracted using a pre-trained visual encoder, projected into the text latent space of an LLM, and subsequently processed by the pre-trained LLM to generate responses. 
Video-ChatGPT~\cite{maaz2024videochatgptdetailedvideounderstanding} introduces instruction tuning into the video domain. Video-LLaVA~\cite{lin2023video} improves model performance with better text-aligned video features \cite{zhu2023languagebind}, while VideoChat2~\cite{li2023mvbench} resorts to increasing the quality and quantity of the video instruction tuning set. Recent vision-LLM models like LLaVA-NeXT~\cite{liu2024llavanext}, LLaVA-OneVision~\cite{li2024llava}, LLaVA-Video~\cite{zhang2024videoinstructiontuningsynthetic}, Qwen2-VL~\cite{Qwen2VL}, and mPlug-Owl3~\cite{ye2024mplugowl3longimagesequenceunderstanding} consider multi-stage training with both video and image, which substantially improves the model performance.
Recent works in VideoLLM focus on long video understanding. ~\cite{wang2024videollamblongcontextvideounderstanding, faure2024hermestemporalcoherentlongformunderstanding, weng2024longvlmefficientlongvideo, korbar2024textconditionedresamplerlongform, zhang2024longcontexttransferlanguage, shen2024longvu} propose to use Q-former~\cite{li2023blip2bootstrappinglanguageimagepretraining} or text-query-based cross-attention to compress vision tokens, while others~\cite{nguyen2024encodingcontrollingglobalsemantics, wang2024longllavascalingmultimodalllms} resort to state-space models~\cite{gu2024mambalineartimesequencemodeling}.
Researchers also extend the instructional tuning into different video sub-domains. For instance, CAT~\cite{ye2024catenhancingmultimodallarge} focuses on audio-visual understanding, while Scene-LLM~\cite{fu2024scenellmextendinglanguagemodel} and LLaVA-3D~\cite{zhu2024llava3dsimpleeffectivepathway} address 3D QA tasks.

Built on these developments, our work specifically focuses on adapting pre-trained Video LLMs to downstream tasks with side-channel signals, aiming to significantly extend the capabilities of these models. 



\medskip
\noindent \textbf{Adaptation of vision foundation models}. Adapting vision foundation models to downstream tasks has received significant attention. Prior works have studied learning lightweight adapters~\cite{rebuffi2017learning,lian2022scaling,bhattacharjee2023vision,chen2022adaptformer,sung2022vl,chen2023vision}, prepending learnable input tokens (\eg prompts)~\cite{zhou2022learning,jia2022visual}, or in-context learning~\cite{wang2023images,xu2024towards,bar2022visual,zhang2023makes}. Recently, adapter- and prompt-based methods have been explored for Video LLMs. Adapt2Reward~\cite{yang2024adapt2reward} re-purposes a video-language model for robotic operation by using it as a language-conditioned reward function. Similarly, SeViLA~\cite{yu2024self} adapts an image-language model for video tasks by introducing Localizer and Answerer modules, derived from BLIP-2~\cite{li2023blip2bootstrappinglanguageimagepretraining}, to enable video event localization and question answering.

Our work is inspired by the success of adapter-based methods such as LoRA~\cite{hu2021loralowrankadaptationlarge}. These methods learn parameter-efficient modules without changing the base model's architecture and weights, and has been widely used to customize large, pre-trained diffusion models for image generation~\cite{kumari2022multiconcept,xiao2023fastcomposer,gu2023mixofshow,wang2023autostory,shah2023ziplora,zhong2024multi}, extending their capabilities in generating images with different styles or controlling the generated content. Moving beyond diffusion models, our goal is to offer a flexible framework that adapts Video LLMs to a wide range of tasks with patch-like adapters, allowing these models to effectively account for additional side-channel signals and adapt to downstream tasks.



\medskip
\noindent \textbf{Multimodal video representation learning}.
Learning a unified representation to connect video and other modalities, such as audio, text and point cloud, has received considerable attention. Prior methods~\cite{luo2022clip4clip} build on the idea of modality alignment using contrastive learning~\cite{radford2021learningtransferablevisualmodels}. More recent works~\cite{chen2023valor, chen2023vast,zhang2023meta,girdhar2023imagebind,zhu2023languagebind} extend the alignment across multiple modalities. Inspired by these works, our method leverages the shared representation between video and text learned by a pre-trained Video LLM, and further embeds side-channel signals into this latent space during adaptation. 

\begin{figure*}[t]
    \centering
    \includegraphics[width=0.85\linewidth]{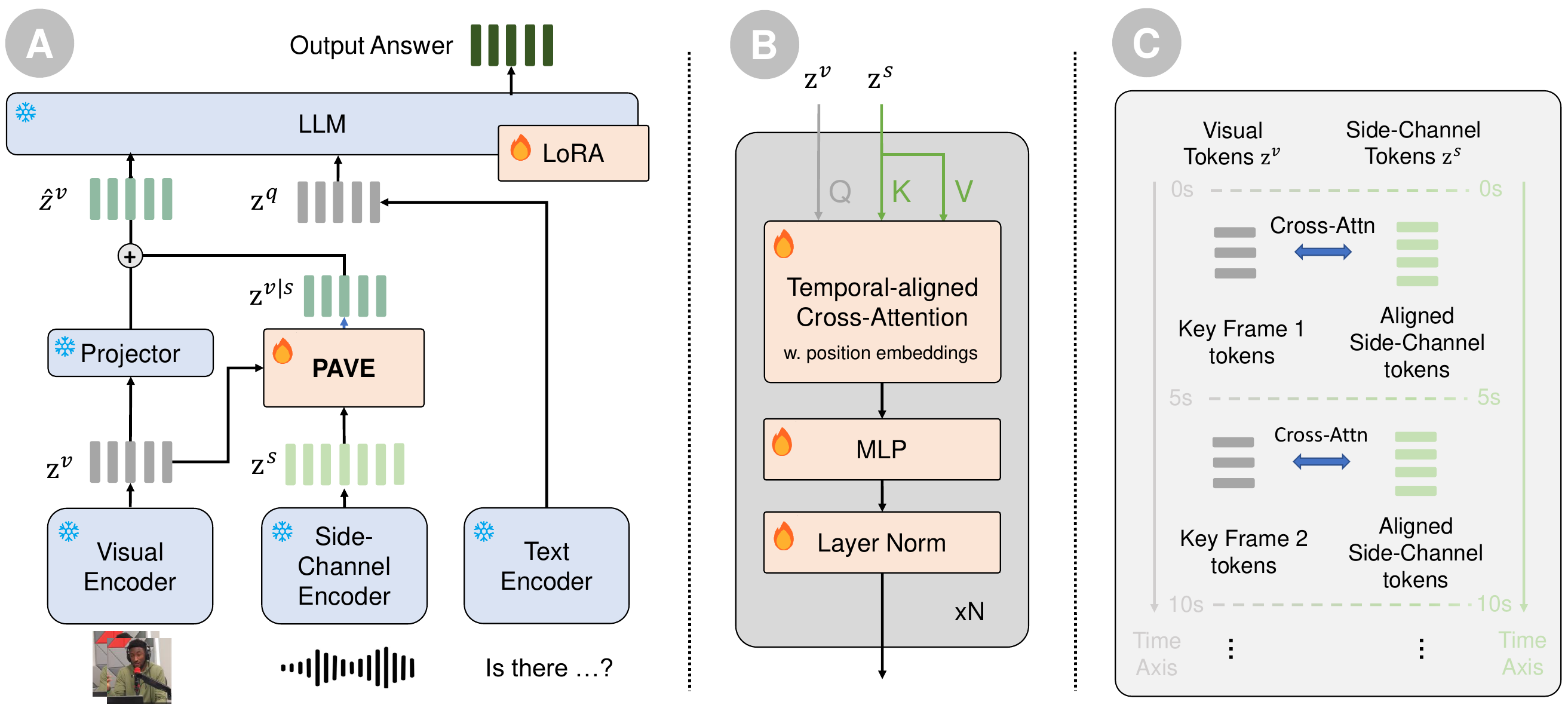}
    \vspace{-0.5em}
    \caption{(a) \textbf{Overview of PAVE}. PAVE presents a simple, parameter-efficient adapter to integrate videos and side-channel signals. This is done by fusing side-channel tokens $\mathbf{z}^s$ and video tokens $\mathbf{z}^v$, and further adding the results to the original video tokens $\mathbf{z}^v$.  (b) \textbf{Details of PAVE's fusion function}. The fusion function $g(\cdot)$ consists of a few blocks of temporal-aligned cross-attention layer, MLP, and layer normalization. (c) \textbf{Temporal-aligned Cross-Attention}. Visual tokens $\mathbf{z}^v$ and side-channel tokens $\mathbf{z}^s$ are aligned along the temporal axis. A video token $\mathbf{z}^v(k)$ is treated as query, and only attends to keys and values (defined by side-channel tokens) in its temporal neighborhood.}
    \label{fig:structure_overview}
    \vspace{-1em}
\end{figure*}

\section{Patching and Adapting Video LLMs}
\label{sec:method}

We propose \textbf{PAVE}, a framework for adapting pre-trained Video LLMs to downstream tasks with side-channel signals. The key to our solution lies in a parameter-efficient, lightweight adapter, referred to as a patch. 
This patch fuses side-channel signals with the original visual tokens and updates them (see Figure~\ref{fig:structure_overview} (a)), and adds a LoRA module~\cite{hu2021loralowrankadaptationlarge} to the LLM, enabling efficient adaptation with a small number of parameters and operations and without altering the base model. 
In what follows, we introduce the background on Video LLM, outline our problem formulation, and present our approach.


\subsection{Preliminaries: Video LLM}
A Video LLM takes a video $\mathbf{X}^v$ and a text query $\mathbf{X}^q=\{x^q\}$ as input, then generates a text answer $\mathbf{X}^a=\{x^a\}$. We assume that $\mathbf{X}^v = \{\mathbf{X}^v_1, \mathbf{X}^v_2, ..., \mathbf{X}^v_K\}$, \ie, a video is represented by $K$ key frames, where $K$ may vary across videos. $\mathbf{X}^v$ is first encoded by a visual encoder
$h_v(\cdot)$ into a set of visual tokens $\{\mathbf{z}^v(k) \in \mathbb{R}^{M \times d}\}$, where $\mathbf{z}^v(k)$ indicates $M$ tokens encoded from the $k$-th key frame within the video ($k \in [1, K]$). Similarly, $\mathbf{X}^q$ is processed by a text encoder $h_t(\cdot)$, which embeds individual words $x^q$ into text tokens $\{\mathbf{z}^q \in \mathbb{R}^d\}$ with $\mathbf{z}^q = h_t(x^q) $. 
These tokens are further combined and processed by an LLM $f(\cdot)$ that decodes $\mathbf{X}^a$ in an autoregressive manner
\begin{equation}
    f\left( \left[ \{\mathbf{z}^v(k)\}, \{\mathbf{z}^q\}, \{\mathbf{z}^a_{<i}\} \right]; \mathbf{\theta} \right) \rightarrow x_i^a, \label{eq:llm}
\end{equation}
where $\{\mathbf{z}^a_{<i}\}$ are text tokens from previously generated answer $x^a_{<i}$, \ie, $\mathbf{z}^a = h_t(x^a)$. $\mathbf{\theta}$ denotes LLM parameters. 

\subsection{Video Tasks with Side-Channel Signals}
We consider adapting a pre-trained Video LLM to downstream tasks with side-channel signals $\mathbf{X}^s$. Similar to videos, we assume that $\mathbf{X}^s = \{\mathbf{X}^s_1, \mathbf{X}^s_2, ..., \mathbf{X}^s_{K_s}\}$, \ie, the side-channel signals also follow a temporal order and are split into $K_s$ blocks. We additionally assume a separate encoder (with possible projector) $h_s(\cdot)$ to process $\mathbf{X}^s$ into a collection of tokens $\{\mathbf{z}^s(k_s) \in \mathbb{R}^{M' \times d} \}$, where $\mathbf{z}^s(k_s)$ is $M'$ tokens encoded from the $k_s$-th block of the side-channel.  It is worth noting that this formulation encapsulates a range of video tasks. We describe some of these tasks considered in our experiments. 

\begin{itemize}
    \item \textbf{Audio-visual understanding}. This task requires jointly processing video and its accompanying audio data (as side-channel signals) for scene understanding, such as event recognition~\cite{xiao2020audiovisual} or QA~\cite{alamri2019audiovisualsceneawaredialog}. 
    \item \textbf{3D scene understanding}. This task focuses on reasoning about the 3D scene using a video of a 3D scan~\cite{azuma_2022_CVPR}. Camera trajectories, as well as an optional sequence of depth maps, constitute the side-channel signals.  
    \item \textbf{Multi-view video understanding}. This task involves combining multi-view videos for visual recognition, \eg, using exocentric videos to complement egocentric video for understanding human activities~\cite{grauman2024ego}. 
    \item \textbf{Enhancing video representations}. This task seeks to enhance the visual representation in the Video LLM. This is done by integrating low resolution, high frame rate frames with the original input of high resolution, low frame rate frames, following the key idea of SlowFast networks~\cite{feichtenhofer2019slowfastnetworksvideorecognition}.

\end{itemize}

\subsection{Our Design of PAVE}

Our goal is to integrate side-channel signals $\mathbf{X}^s$ into the LLM to enhance video reasoning, without modifying the structure of $f(\cdot)$ and encoders ($h_v(\cdot)$ and $h_t(\cdot)$), nor adding any major set of parameters. To achieve this goal, our key idea is learning a function $g(\cdot)$ to fuse side-channel tokens $\{\mathbf{z}^s(k_s)\}$ with the original visual tokens $\{\mathbf{z}^v(k)\}$. The fusion results have the same size of $\{\mathbf{z}^v(k)\}$, and will be further used to update $\{\mathbf{z}^v(k)\}$. Formally, our design of PAVE, as shown in Figure~\ref{fig:structure_overview} (a), is expressed as
\begin{equation}
\begin{split}
    \text{fusion:} \quad & \{\mathbf{z}^{v|s}(k)\} = g([\{\mathbf{z}^s(k_s)\}, \{\mathbf{z}^v(k)\}]; \phi)\\
    \text{summation:} \quad & \mathbf{\hat{z}}^{v}(k) = \mathbf{z}^{v}(k) + \mathbf{z}^{v|s}(k),
\end{split}\label{eq:pave}
\end{equation}
where $\phi$ denotes learnable parameters of $g(\cdot)$. 

Intuitively, $g(\cdot)$ injects side-channel information into a set of tokens of the same size as the original visual tokens, based on which a simple summation can be performed to form a residual connection. A key feature of this design is that the number of input tokens to $f(\cdot)$ remains unchanged. As the main computational cost lies in $f(\cdot)$, doing so results in a negligible overhead. 

\medskip
\noindent \textbf{The fusion function $g(\cdot)$}. Our fusion function is realized using a variant of cross-attention, as illustrated in Figure~\ref{fig:structure_overview} (b). Specifically, the vision tokens $\{\mathbf{z}^v(k)\}$ are transformed into the queries, and the side-channel tokens $\{\mathbf{z}^s(k_s)\}$ form the keys and values. To align video and side-channel signals in time while maintaining a low computation cost, we consider a local cross-attention, named temporal-aligned cross-attention, where a query $\mathbf{z}^v(k)$ only attends to keys and values in its temporal neighborhood, \ie, $\{\mathbf{z}^s(k_s)\}$ with $k_s \in N(k)$ (see Figure~\ref{fig:structure_overview} (c)). Before computing the cross attention, rotary positional embeddings~\cite{su2024roformer} are added to the queries and keys. After cross attention, an MLP with layer normalization is applied to further transform the features, similar to a standard Transformer block~\cite{vaswani2017attention}.

\medskip
\noindent \textbf{Integration with LoRA and training loss}. We further combine PAVE with LoRA~\cite{hu2021loralowrankadaptationlarge} by adding a small set of parameters in the form of low rank approximation $\Delta \theta$ to the LLM $f(\cdot)$. Putting things together, PAVE minimizes the standard negative log likelihood of its output when adapting to a downstream task. This is given by 
\begin{equation*}
\argmin_{\mathbf{\Delta\theta}, \mathbf{\phi}} \ \mathbb{E}_{\mathcal{D}} \ \left[ -\log p \left( x^a_i | \left[ \{\mathbf{\hat{z}}^{v}\}, \{\mathbf{z}^{q}\}, \{\mathbf{z}^a_{<i}\} \right]; \theta+\Delta \theta, \phi \right) \right],
\end{equation*}
where $\mathcal{D}$ is the data distribution approximated by the training set $(\mathbf{X}^v, \mathbf{X}^q, \mathbf{X}^a) \sim \mathcal{D}$. $\{\mathbf{\hat{z}}^{v}\}$ is computed using Eq.\ \ref{eq:pave}, thus creating a dependency on $g(\cdot)$ and its parameters $\phi$.
\section{Experiments and Results}\label{sec:experiments}
Our main experiments include (1) audio-visual QA (Sec.\ \ref{section_res_audio}), (2) 3D QA (Sec.\ \ref{section_res_3d}, (3) enhancing video QA by considering high frame rate videos (Sec\ \ref{section_res_video}), and (4) multi-view video recognition (Supplement). In addition, we ablate our design, investigate cross-model generalization, and demonstrate multi-task joint training in Sec.\ \ref{section_res_ablation}. Further implementation details for individual experiments can be found in the Supplement.  


\begin{table*}[t]  
\centering  
\scalebox{0.7}{
\begin{tabular}{lc|c|cccc|c|c}  
\toprule
       \multirow{2}{*}{Method}      & AVSD~\cite{alamri2019audiovisualsceneawaredialog}  &AVQA~\cite{yang2022avqa}&\multicolumn{4}{c|}{MUSIC-AVQA~\cite{Li2022Learning}}  & \multirow{2}{*}{TFLOPs} & \multirow{2}{*}{Total / Trainable Params} \\
                                    & CIDEr                                              & Acc. &Audio Acc. & Visual Acc. & Audio-Visual Acc.  & Overall Acc. & & \\
    \midrule
    \multicolumn{3}{l}{\it \small\textbf{Zero-shot LMMs} } \\
     CAT-7B ~\cite{ye2024catenhancingmultimodallarge}             & 79.0 & -  & -  & -    & -    & 48.6 & -      & - \\
     LLaVA-OV-7B~\cite{li2024llava}                               & 70.6 & 85.6  & \textcolor{gray}{68.8} & 70.6 & 52.8 & 60.4 & 98.53 & 8.2B~/~-\\
     \midrule
     \multicolumn{3}{l}{\it \small\textbf{Task-specific models} } \\     
     MTN~\cite{Le_2019}                                           & 98.5  & -  & - & - & -  &  - &-  & -\\
     COST~\cite{pham2022videodialogconversationobjects}           & 108.5 & -  & - & - & -  &  - &-  & - \\
     VALOR~\cite{chen2023valor} & -  & - & - & -  & -&  78.9 & - & - \\
     VAST~\cite{chen2023vast} & -  & - & - & -  & -&  80.7 & - & - \\
     PSTP-Net~\cite{li2023progressive} & -  & 90.2 & - & -  & -&  - & - & - \\
     CAT-7B-FT ~\cite{ye2024catenhancingmultimodallarge}          & - & 92.0 & \textcolor{gray}{\textbf{84.9}} & 86.1 & \textbf{83.2} &    \textbf{84.3} & -  & - \\ 
     LLaVA-OV-7B-FT                                               & 124.9 & 90.8 & \textcolor{gray}{75.4} & 89.3 & 72.3 &  77.4  &98.53 & 8.2B~/~161.5M \\
     \midrule
     
     PAVE-7B   (w/ audio)                                          & \textbf{152.9} & \textbf{93.8} & \textcolor{gray}{79.7} & \textbf{93.0} & 78.0 &  82.3 & 98.63 & 8.2B~/~170.5M \\ \bottomrule

\end{tabular}
}
\vspace{-1mm}
\caption{\textbf{Results of audio-visual QA} with audio as side-channel signals. We report CIDEr scores on AVSD and the accuracy (Acc.) on AVQA and Music-AVQA. LLaVA-OV-7B-FT refers to directly fine-tuning the LLaVA-OneVision using video. \textcolor{gray}{Gray} indicates results that require audio-inference only. Our model achieves state-of-the-art performance on AVSD, AVQA and visual split of Music-AVQA by only adding a small amount of parameters and FLOPs.
\vspace{-4mm}
}  
\label{tab:video_audio_understanding}  
\end{table*}  

\subsection{Results on Audio-Visual QA} \label{section_res_audio}

Audio-visual QA aims to answer questions related to various visual and auditory concepts (\eg, objects and the sound they produce), as well as their interactions in videos, based on an input of video with its accompanying audio. In this task, we treat audio as the side-channel signals. We now describe our experiment protocol, implementation details, baselines, and results. 



\medskip
\noindent\textbf{Experiment protocol.} Our experiments consider both open-end QA (AVSD~\cite{alamri2019audiovisualsceneawaredialog}) and closed-end QA (AVQA~\cite{yang2022avqa} and Music-AVQA~\cite{Li2022Learning}). We follow the protocol in previous works~\cite{ye2024catenhancingmultimodallarge, pham2022videodialogconversationobjects} to train and evaluate PAVE using corresponding splits of the datasets. For the open-end AVSD, we use the AVSD@DSTC7 test split and report the CIDEr score as the metric. For the closed-end AVQA and Music-AVQA, we evaluate PAVE on the standard eval / test split, and report the accuracy as the metric. For PAVE, we additionally calculate the model inference FLOPs during prefilling, as well as the total / trainable number of parameters. 


\medskip
\noindent\textbf{Implementation details.} 
To encode audio we extract audio features using ImageBind~\cite{girdhar2023imagebind}. It samples the audio at a 16kHz rate, generating approximately 1 audio token per second. PAVE further integrate these audio tokens into a Video LLM (LLaVA-OneVision~\cite{li2024llava}). We follow the LLaVA-OneVision~\cite{li2024llava} to evenly sample 32 video frames from an input video.
We train PAVE 1 epoch with the AVSD / 2 epochs with AVQA / 2 epochs with Music-AVQA training set, respectively. We then evaluate PAVE's performance on their corresponding test set. 

\medskip
\noindent\textbf{Baselines.}
We consider two types of baseline methods: 
(1) Multimodal LLMs designed for general video or audio-visual understanding, \eg, LLaVA-OneVision~\cite{li2024llava} and CAT~\cite{ye2024catenhancingmultimodallarge}, which have never been trained on the target dataset (\ie, zero-short inference); and 
(2) Task-specific models, which has been fine-tuned on the target dataset, such as MTN~\cite{Le_2019} and COST~\cite{pham2022videodialogconversationobjects} for AVSD, PSTP-Net~\cite{li2023progressive} and CAT-7B-FT~\cite{ye2024catenhancingmultimodallarge} for AVQA, and the VALOR~\cite{chen2023valor}, VAST~\cite{chen2023vast} and CAT-7B-FT for Music-AVQA. 
We also include a baseline that directly fine-tunes our base model (LLaVA-OneVision) with LoRA on the training set without using the audio, denoted as LLaVA-OV-7B-FT. 

\begin{table*}[t]  
\centering  
\scalebox{0.75}{
\begin{tabular}{lccccc|c|c|c}  
\toprule
\multirow{2}{*}{Method}                                          & \multicolumn{5}{c|}{ScanQA~\cite{azuma_2022_CVPR}} & SQA3D~\cite{ma2022sqa3d} & \multirow{2}{*}{TFLOPs} & \multirow{2}{*}{Total / Trainable Params}\\
                                                                 & C     & B-4 & M     & R    & EM@1 & EM@1     &&            \\ 
    \midrule
    \multicolumn{6}{l}{\it \small\textbf{Zero-shot LMMs}} \\
     VideoChat2-7B~\cite{li2024mvbenchcomprehensivemultimodalvideo} & 49.2  & 9.6 & 9.5   & 28.2 & 19.2 & 37.3 & - & - \\
     LLaVA-OV-7B~\cite{li2024llava}                              & 91.0  & 5.3 &  18.2 & 45.9 & 26.7 \textcolor{gray}{(44.3)} & 8.3 \textcolor{gray}{(50.7)} &98.53 & 8.2B / - \\
    \midrule
    \multicolumn{6}{l}{\it \small\textbf{Task-specific models}} \\
     3D-LLM-7B~\cite{hong20233d} & 74.5 & 12.9 & 15.1 & 37.5 & 21.2 & 49.79 &- & - \\
     LEO-7B~\cite{huang2023embodied} & 101.4 & 13.2 & 20.0 & 49.2 & 24.5  \textcolor{gray}{(47.6)} & 50.0 \textcolor{gray}{(52.4)} & - & -  \\
     Scene-LLM-7B~\cite{fu2024scenellmextendinglanguagemodel}       &  80.0 & 12.0 & 16.6 & 40.0 & 27.2 & 54.2 & - & - \\
     LLaVA-3D-7B~\cite{zhu2024llava3dsimpleeffectivepathway}        &  91.7  & 14.5 & \textbf{20.7} & \textbf{50.1} & 27.0 \textcolor{gray}{(45.0)} & 55.6 \textcolor{gray}{(57.6)} & - & - \\
     
      LLaVA-OV-7B-FT                                                         & 95.1 & 13.5 & 19.1 & 47.4 & 27.4 \textcolor{gray}{(46.3)} & 55.8 \textcolor{gray}{(58.1)} & 98.53 & 8.2B / 161.5M \\
    \midrule
     
     PAVE-7B   (w/ 3D info)                                       & \textbf{103.4} & \textbf{16.0} & 19.9 & 49.0 & \textbf{29.1} \textcolor{gray}{(48.5)} & \textbf{59.0} \textcolor{gray}{(61.4)} & 98.68 & 8.2B / 170.5M\\ \bottomrule

\end{tabular}
}
\vspace{-1mm}
\caption{\textbf{Results of 3D QA} with 3D cues as side-channel signals. We report CIDEr (C), BLEU-4 (B-4), METEOR (M), ROUGE(R) for ScanQA, and top-1 Exact Match (EM@1) for both ScanQA and SQA3D datasets. LLaVA-OV-7B-FT refers to directly fine-tuning the LLaVA-OneVision on ScanQA or SQA3D. \textcolor{gray}{Gray} indicates evaluation results with refined exact-match protocol. PAVE achieves state-of-the-art performance on both ScanQA and SQA3D with a small number of additional parameters and FLOPs.
}  
\vspace{-3mm}
\label{tab:3d_qa_understanding}  
\end{table*}  

\medskip
\noindent\textbf{Results and discussion.}
Table~\ref{tab:video_audio_understanding} summarizes the results. PAVE outperforms COST~\cite{pham2022videodialogconversationobjects} by 44 points in CIDEr scores on AVSD and surpasses CAT-7B-FT by $\sim$2\% on AVQA, achieving state-of-the-art results. 
On Music-AVQA, PAVE outperforms latest methods by 7\% in the visual split, where the answers can be derived from input video, but lags behind CAT-7B-FT on audio and audio-visual splits, where the answers are from audio alone or must be reasoned by jointly considering audio and video. Compared to LLaVA-OV-7B-FT, PAVE consistently improves performance across three datasets, adding only 9M parameters and 0.1 TFLOPs ($\sim$0.1\% of total parameters and FLOPs). 

While PAVE is not designed for audio-only inference (Music-AVQA's audio split), its performance gap on the audio-visual split is puzzling. Our observation is that a part of the questions in this audio-visual split deliberately feature conflicting audio and visual information, \eg, the sound of a piano coupled with the video with a guitar (see the last example in Figure~\ref{fig:qa_visualization}). We find that PAVE, based on a Video LLM trained on video-text data, often prioritizes visual cues, leading to degraded performance.


\subsection{Results on 3D QA} \label{section_res_3d}
3D QA focuses on the reasoning about spatial location of individual objects or relative position between objects, based on an input video scan with camera poses and depth information. Previously, the input data is often converted into a voxel or mesh-based 3D representation, based on which the reasoning is performed. In this work, we instead treat the 3D cues (camera poses and depth maps) as the side-channel, and leverage pre-trained Video LLMs for 3D QA. Again, we describe our experiment protocol, implementation details, baselines, and results.

\medskip
\noindent\textbf{Experiment protocol.} Our experiments consider ScanQA~\cite{azuma_2022_CVPR} and SQA3D~\cite{ma2022sqa3d} datasets, using their corresponding train / test splits. 
Following previous work~\cite{zhu2024llava3dsimpleeffectivepathway}, we report the CIDEr (C), BLEU-4 (B-4), METEOR (M), ROUGE(R), and top-1 Exact Match (EM@1) metrics on ScanQA and report EM@1 on SQA3D. Similarly, we also report model inference FLOPs and parameters for PAVE. 

\medskip
\noindent\textbf{Implementation details.}
To encode 3D cues, we use the 3D encoder from LLaVA-3D~\cite{zhu2024llava3dsimpleeffectivepathway}. It encodes camera pose, RGB images, and depth information into multi-view features defined on the RGB image plane. We again build PAVE with LLaVA-OneVision~\cite{li2024llava}. Specifically, we evenly sample 32 RGB-D frames with their camera poses from the scanning. The 3D encoder creates a sequence of 2D feature maps (\ie, multi-view features), leading to around 18K 3D tokens per scan. The visual encoder in LLaVA-OneVision separately embeds the 32 key RGB frames into video tokens. PAVE further integrates video tokens and 3D tokens. We train the PAVE 1 epoch with ScanQA / 2 epochs with SQA3D training set, and then evaluating PAVE performance on their test sets, respectively.

\medskip
\noindent\textbf{Baselines.} We again compare our method with two types of baselines: 
(1) Video LLMs for general video understanding, \eg, LLaVA-OneVision\cite{li2024llava} and VideoChat2~\cite{li2023mvbench}, with zero-shot inference; and 
(2) Task-specific models fine-tuned on the target dataset, \eg, LEO~\cite{huang2023embodied}, 3D-LLM~\cite{hong20233d}, Scene-LLM~\cite{fu2024scenellmextendinglanguagemodel} and LLaVA-3D~\cite{zhu2024llava3dsimpleeffectivepathway}. 
We also add a baseline that directly fine-tunes LLaVA-OneVision with LoRA without using 3D cues, denoted as LLaVA-OV-7B-FT. 

\medskip
\noindent\textbf{Results.} Our results are presented in Table~\ref{tab:3d_qa_understanding}. In comparison to our baselines, PAVE achieves highly competitive performance across ScanQA and SQA3D datasets. In particular, PAVE attains significant gains in CIDEr and EM@1 scores. For example, on EM@1, PAVE outperforms the best reporting result by 2-3\% in absolute values. Importantly, PAVE only adds 9M parameters and 0.15 TFLOPs on top of the base model during inference, accounting for $\sim$0.1\% of the total parameters and compute cost. Interestingly, fine-tuning the base model using videos only and without 3D cues (LLaVA-OV-7B-FT) also has competitive results, indicating strong capability of Video LLMs for 3D reasoning.

\subsection{Results on Enhanced Video QA} \label{section_res_video}

\begin{table*}[t]  
\centering  
\scalebox{0.75}{
\begin{tabular}{l|cccc|cccc|c|c|c}  
\toprule
\multirow{2}{*}{Method}   & \multicolumn{4}{c|}{VideoMME~\cite{fu2024video}} & \multicolumn{4}{c|}{MVBench~\cite{li2023mvbench}} & \multirow{2}{*}{MLVU~\cite{zhou2025mlvubenchmarkingmultitasklong}}  & \multirow{2}{*}{FLOPs(TB)} & \multirow{2}{*}{Total / Trainable Params}  \\
                                                 & Short   & Median   & Long    & Avg   & SC  & FGP & OS & Avg                   & &                             & \\ 
    \midrule
     LLaVA-OV-0.5B~\cite{li2024llava}     & 53.4    & 41.2     & 37.3    & 44.0  & 37.5  &  49.0 & 33.0 & 45.5                   & 50.3 & 8.01 & 0.9B~/~- \\
     LLaVA-OV-7B~\cite{li2024llava}       & 70.1    & 56.6     & 48.9    & 58.2  & \textbf{52.0} &  53.0& 35.5 & 56.7            & 64.7 & 98.53 & 8.2B~/~- \\
     \midrule
     PAVE-0.5B (w/ video feature)        & \textbf{57.8} & \textbf{42.7} & \textbf{37.4} & \textbf{46.0} & \textbf{40.0} & \textbf{54.0} & \textbf{33.5} & \textbf{46.6} &  \textbf{51.6} & 8.08  & 0.9B~/~41.4M  \\
     PAVE-7B   (w/ video feature)          & \textbf{71.1} & \textbf{59.4} & \textbf{49.2} & \textbf{59.9} & 51.5   & \textbf{54.5} & \textbf{39.0}      & \textbf{58.0} & \textbf{67.0}& 98.63 & 8.2B~/~170.5M \\ \bottomrule

\end{tabular}
}
\vspace{-1mm}
\caption{\textbf{Results of using high frame rate videos to enhance video QA}, where densely sampled video frames are treated as side-channel signals. 
For MVBench, we report performance on state change (SC), fine-grained pose (FGP), and object shuffle (OS) subsets. 
PAVE consistently outperforms baselines on MLVU, VideoMME, and the tasks that need fine-grained motion information in MVBench.
\vspace{-4mm}
}  
\label{tab:general_video_understanding}  
\end{table*}  

Video LLMs~\cite{li2024llava,lin2023video} often sample a fixed number of frames (\eg 8 or 32) to represent the input video. These sparsely sampled frames may miss detailed temporal information, especially for tasks that require fine-grained motion information. We now consider injecting high frame rate videos as the side-channel signals. We describe our experiment protocol, implementation details, baselines, and results.  


\medskip
\noindent\textbf{Experiment protocol.} To simply our experiment, we create a subset from LLaVA-Video-178K \cite{zhang2024videoinstructiontuningsynthetic} for training PAVE. We first sample all videos longer than 1 minute and then randomly choosing 2 question-answer pairs for each video, leading to 57K videos and 114K question-and-answer pairs. To evaluate PAVE, we consider a set of video benchmarks including VideoMME~\cite{fu2024video}, MVBench~\cite{li2023mvbench}, and MLVU~\cite{zhou2025mlvubenchmarkingmultitasklong}. VideoMME and MVBench are both comprehensive video benchmarks and cover different types of subtasks, while MLVU focuses on long video understanding.
All benchmarks use accuracy as the performance metric. Similar to other experiments, we report inference FLOPs and parameters for PAVE. Results on additional benchmarks are included in the Supplement.

\medskip
\noindent\textbf{Implementation details.} For side-channel signals, we densely sample the video frame at 2 fps and use the LanguageBind~\cite{zhu2023languagebind} to extract features. Motivated by SlowFast networks~\cite{feichtenhofer2019slowfastnetworksvideorecognition}, we downsample the spatial resolution of LanguageBind's features to $2 \times 2$ and treat it as the fast stream, while the original visual features used by the base Video LLM (LLaVA-OneVision) are regarded as the slow stream. 
We 
train PAVE for 1 epoch using our subset.

\medskip
\noindent\textbf{Baselines.}
We use the LLaVA-OneVision 0.5B and 7B models as our baselines, as they achieve top performance on various video benchmarks. This setup allows us to assess PAVE's performance across different model sizes.

\medskip
\noindent\textbf{Results.} Table~\ref{tab:general_video_understanding} presents our results. In comparison to the base model (LLaVA-OneVision), PAVE achieves consistent improvements across 3 datasets and 2 model sizes. 
On VideoMME, PAVE achieves major boosts (1.0-4.4\%) on the short and median splits, which contain videos below 15 minutes on average, and only minor gains (0.1-0.3\%) on longer videos ranging from 30 minutes to 1 hour. 
On MVBench, PAVE improve the baseline by 1.1\% to 1.3\% on average. Specifically, PAVE-0.5B outperforms the baseline by 5\% on the fine-grained pose (FGP) task, and PAVE-7B improves by 4.5\% on the object shuffle (OS) task.
On the long video benchmark MLVU, PAVE shows a significant improvement (1.3-2.3\%).


\subsection{Further Analysis} \label{section_res_ablation}
\noindent\textbf{Ablation: design of PAVE.}
In this part, we conduct an ablation study to analyze key design choices of PAVE. 
We ablate two possible approaches to combine signal-channel tokens $\mathbf{z}^s$ with the original video tokens $\mathbf{z}^v$: (1) Interleaved with the $\mathbf{z}^v$ as used in~\cite{li2024llavainterleave}, noted as FT (interleave), resulting in significantly heightened computational cost due to the increased number of input tokens. 
(2) Added to $\mathbf{z}^v$ as considered in PAVE, where the summation operation requires that side-channel tokens $\mathbf{z}^s$ have the same shape as video tokens $\mathbf{z}^v$. 
When using the summation operation, we further consider two design choices for the cross-attention: (a) Considering learnable tokens as the query, similar to~\cite{jaegle2022perceiveriogeneralarchitecture}, noted as the PAVE (learnable) and 
(b) Using the original visual tokens $\mathbf{z}^v$ as the query, noted as the PAVE (visual). 
For reference, we also include two baselines: (Zero-shot) the zero-shot performance of Video LLM on the ScanQA, and (FT) Video LLM performance on the ScanQA after being fine-tuned on the ScanQA training split with LoRA.


We conduct experiments with the LLaVA-OneVision 0.5B model on the ScanQA dataset. We report top-1 Exact Match (EM@1), FLOPs and trainable parameters
%
in Table~\ref{tab:adptor_design}. PAVE (visual) achieves the best performance. Compared with FT (interleave), PAVE effectively reduces the number of tokens sent to the LLM, thereby drastically reducing FLOPs. Further, PAVE (visual) outperforms FT (interleave) in both accuracy and efficiency, indicating that it can more effectively incorporate side-channel information.
For query design, we observe that PAVE (visual) yields better performance than PAVE (learnable). We conjecture that explicit alignment between video tokens and side-channel tokens may be critical here. 


\begin{table}[t]  
\centering  
\scalebox{0.8}{
\begin{tabular}{lccc}  
\toprule
       Design                  & EM@1 & FLOPs (TB) &  Trainable Params \\
     \midrule
     Zero-shot                 &  0.2 \textcolor{gray}{(28.0)} & 8.01  & -    \\
     FT                      & 20.5 \textcolor{gray}{(36.3)} & 8.01  & 35.1M  \\
     FT (interleave)         & 23.0 \textcolor{gray}{(39.9)}  & 71.41 & 35.1M \\
     
     PAVE (learnable)      & 22.5 \textcolor{gray}{(39.3)}  & 8.13 & 43.9M\\
     PAVE (visual)       & \textbf{23.1} \textcolor{gray}{(40.0)}  & 8.13 & 41.4M\\ \bottomrule

\end{tabular}
}
\vspace{-1mm}
\caption{\textbf{Ablation study} on the design of PAVE on 3DQA ScanQA benchmark. We report the Top-1 Exact Match score, FLOPs, and the number of trainable parameters. 
\vspace{-2mm}
}  
\label{tab:adptor_design}  
\end{table}  

\medskip
\noindent\textbf{Generalization across Video LLMs.}
Moving forward, we demonstrate that PAVE can be applied to different models and various model scales.
We train PAVE using pre-trained LLaVA-Onevision~\cite{li2024llava} and Video-LLaVA~\cite{lin2023video} as base models. 
We choose LLaVA-OneVision as it provides models at both 0.5B and 7B scales, and Video-LLaVA since its pre-training set and model structure differ from those of LLaVA-OneVision, potentially impacting PAVE's adaptation. 
We conduct experiments on the ScanQA dataset, reporting top-1 Exact Match (EM@1) as the metric.
Table~\ref{tab:adapt_to_different_model} presents our results. PAVE can adapt Video LLMs at various scales, accommodate different model architectures, and work with models trained on diverse instruction datasets.

\begin{table}[t]  
\centering  
\scalebox{0.75}{
\begin{tabular}{lccc}  
\toprule
       Method                  & LLaVA-OV-0.5B & LLaVA-OV-7B &  Video-LLaVA-7B~\cite{lin2023video} \\
     \midrule
     Zero-shot                 & 0.2 \textcolor{gray}{(28.0)}  & 26.7 \textcolor{gray}{(44.3)} & 0.0 \textcolor{gray}{(25.1)} \\
     Finetune                  & 20.5 \textcolor{gray}{(36.3)} & 27.4 \textcolor{gray}{(46.3)}     & 21.6 \textcolor{gray}{(37.3)}\\
     PAVE                       & 23.1 \textcolor{gray}{(40.0)} & {29.1} \textcolor{gray}{(48.5)}      & 25.2 \textcolor{gray}{(42.1)}\\ \bottomrule

\end{tabular}
}
\vspace{-1mm}
\caption{\textbf{Generalization} of PAVE to Video LLMs with different architectures and sizes. PAVE shows consistent improvements across 3 settings, demonstrating its effectiveness and versatility.
\vspace{-1mm}
}  
\label{tab:adapt_to_different_model}  
\end{table}  


\begin{table}[t]  
\centering  
\scalebox{0.65}{
\begin{tabular}{lccc}  
\toprule
       Model                  & CIDEr & FLOPs (TB) &  Total / Trainable Params \\
     \midrule
     PAVE-7B (audio)                   &  152.5  & 98.53 & 8.2B / 170.3M \\      


     PAVE-7B (audio + dense-frames)    & \textbf{160.0} & 98.74 & 8.2B / 170.3M \\ \bottomrule

\end{tabular}
}
\vspace{-2mm}
\caption{\textbf{Multi-task learning with PAVE} on the AVSD dataset. Combining both audio and high frame rate patches leads to substantial performance improvement.
\vspace{-4mm}
}  
\label{tab:avsd_with_dense_frames}  
\end{table}
\begin{figure*}[t!]
    \centering
    \includegraphics[width=0.92\linewidth]{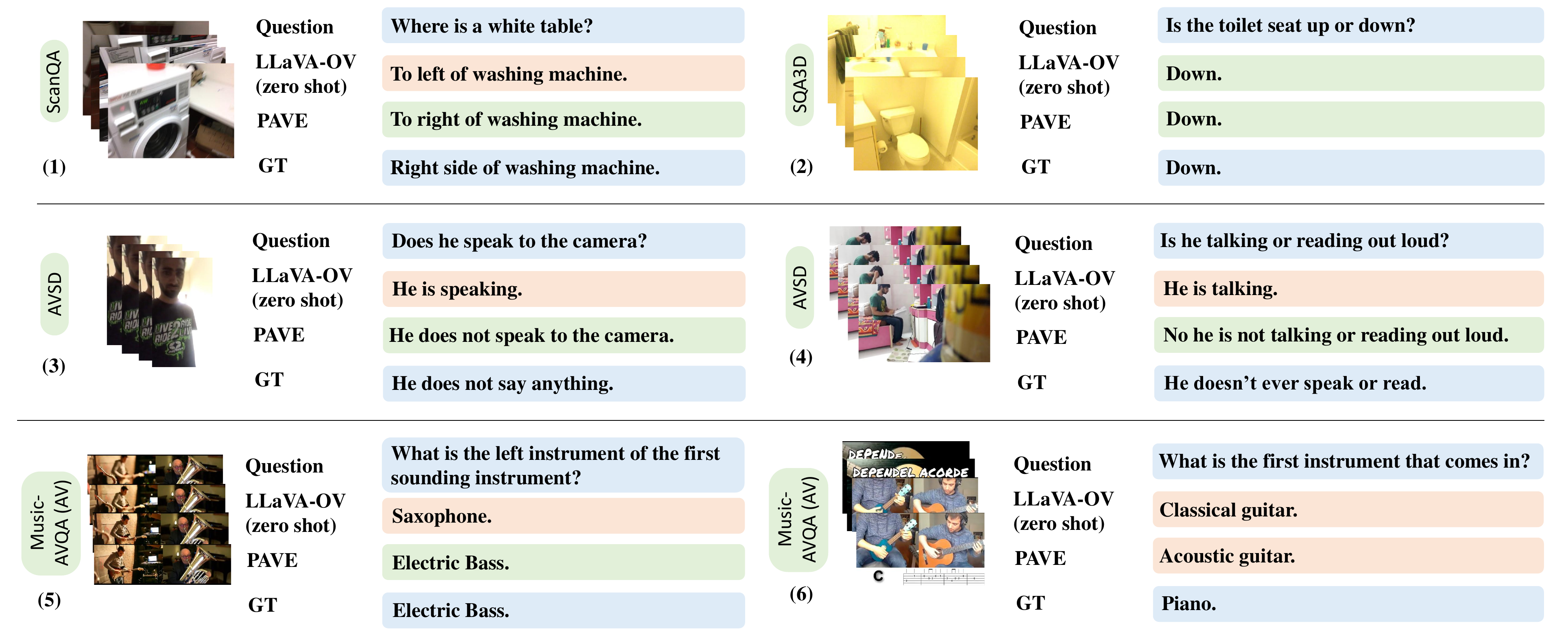}
    \vspace{-1em}
    \caption{\textbf{Visualization of sample results}. We visualize the compare the results from our base model LLaVA-OneVision (under zero-shot inference) and PAVE across 3D QA and audio-visual QA tasks. Both succeful and failure cases are shown. 
    }
    \label{fig:qa_visualization}
    \vspace{-0.5em}
\end{figure*}

\begin{figure*}[t!]
    \centering
    \includegraphics[width=0.90\linewidth]{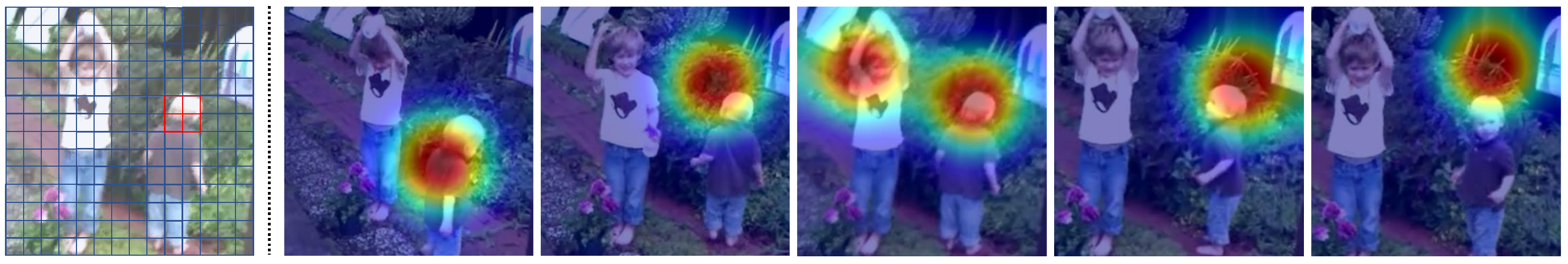}
    \vspace{-0.5em}
    \caption{\textbf{Visualization of cross-attention scores in PAVE} when injecting high frame rate videos as the side-channel. Cross-attention scores are calculated between the selected video tokens from the original low frame rate video (red cells on the left) and side-channel tokens from the high frame rate video. Scores are displayed as heatmaps over densely sampled video frames (on the right).} 
    \label{fig:cross_attention}
    \vspace{-1.5em}
\end{figure*}

\medskip
\noindent \textbf{Multi-task learning.} Moreover, we explore the potential of training multiple patches simultaneously and study the effects of such multi-task learning. We consider the setting of multiple tasks share overlapping side-channels (\eg, high-frame-rate video). In this case, multiple patches can be learned jointly, leading to potential enhancement across their corresponding side-channels.

As a first step to demonstrate this possibility, we design an experiment to integrate high frame rate video patch and audio-visual path for high frame rate audio-visual QA. Specifically, we build on the patch learned for injecting high frame rate videos (PAVE-7B model in Table~\ref{tab:general_video_understanding}), and train an additional patch for audio-visual QA on the AVSD training set. During training, the old patch is kept frozen and only the audio-visual patch is learned.  During the inference, we activate both patches, and PAVE takes input key frames (\ie low frame rate video), high frame rate video, and audio as input. Table~\ref{tab:avsd_with_dense_frames} shows that our multi-task learning leads to a notable improvement (\textit{+7.5} in CIDEr scores) on the AVSD test set, demonstrating the possibility and the benefit of training multiple patches together. We provide further discussion in Sec.\ \ref{sec:conclusion}.




\medskip
\noindent\textbf{Results visualization and diagnosis.} 
Finally, we visualize sample results of 3D QA and audio-visual QA in Figure~\ref{fig:qa_visualization}. An interesting failure case of PAVE, as we previously mentioned in Sec.\ \ref{section_res_audio} is shown in sample (6), where the piano is present in the audio but never appears visually in the video, creating a conflict between the auditory and visual information. In this case, PAVE produces inaccurate answers.

To further diagnose our model, we visualize cross-attention scores between video tokens and side-channel tokes in PAVE's fusion function. Figure~\ref{fig:cross_attention} shows  heatmaps of cross-attention scores when injecting high high frame rate videos as the side-channel (corresponding to results of PAVE-7B in Table~\ref{tab:general_video_understanding}). We calculate the cross-attention scores between the selected video tokens (shown in the red cells) from the original low frame rate videos, and the side-channel tokens from the high frame rate videos. The visualization shows that the peaks of the heatmap keep track of visually similar regions in densely sampled video frames. This result indicates that the fusion function in PAVE facilitates explicit alignment between video tokens and side-channel tokens, and thus allows for effectively integrating information from the side-channel.

\section{Conclusion and Discussion}
\label{sec:conclusion}



In this paper, we addressed the problem of adapting pre-trained Video LLMs to downstream tasks involving side-channel signals --- additional modalities or data types such as audio, 3D cues, high frame rate or multi-view videos.  
To this end, we presented PAVE, a flexible framework that enables adaptation through patching. 
PAVE built on lightweight adapters (\ie, ``patches''), which adds a small number of parameters and operations ($\sim$0.1\%) to a base model without modifying its architecture or pre-trained weights.
We demonstrated that PAVE can effectively adapt various video LLMs across multiple tasks, often surpassing state-of-the-art task-specific models. 
Additionally, we showed encourage results of learning across multiple tasks with PAVE.
We believe that our work provides a solid step towards adapting Video LLMs for diverse tasks, and sheds light on the broader topic of multi-modal reasoning.


\medskip
\noindent \textbf{Distributing and deploying tasks-specific patches.} A key strength of PAVE is that only a small patch is need for adapting a large pre-trained Video LLM to a new task. For example, for LLaVA-OneVision-7B, this patch is only $\sim$20 MB, while the model occupies $\sim$16 GB. We anticipate that this compactness makes it practical to distribute and deploy individual patches, similar to LoRA weights used in diffusion models, thereby facilitating broader adoption of Video LLMs in downstream applications.

\medskip
\noindent \textbf{Towards multi-task learning.} An exciting future research direction is the joint learning of multiple patches across tasks. While we have demonstrated the feasibility of stage-wise multi-task learning, future work should aim to handle multi-task joint training and explore methods for combining individual patches to enable flexible task composition.

\medskip
\noindent \textbf{Acknowledgment}: This work was partially supported by National Science Foundation under Grant No.\ CNS 2333491, by the Army Research Lab under contract number W911NF-2020221, and by a contract from General Motors. 
{
    \small
    \bibliographystyle{ieeenat_fullname}
    \bibliography{main}
}

\clearpage
\appendix
In this supplement, we (1) show additional experiment results on small Video LLM and multiple-view video understanding (Section~\ref{experiment}); (2) describe additional implementation details (Section~\ref{implementation}); (3) include additional visualization of the question-answering results (Section~\ref{visualization}).

\begin{table*}[t]  
\centering  
\scalebox{0.77}{
\begin{tabular}{lc|c|cccc|c|c}  
\toprule
       \multirow{2}{*}{Method}      & AVSD~\cite{alamri2019audiovisualsceneawaredialog}  &AVQA~\cite{yang2022avqa}&\multicolumn{4}{c|}{MUSIC-AVQA~\cite{Li2022Learning}}  & \multirow{2}{*}{TFLOPs} & \multirow{2}{*}{Total / Trainable Params} \\
                                    & CIDEr                                              & Acc. &Audio Acc. & Visual Acc. & Audio-Visual Acc.  & Overall Acc. & & \\
    \midrule
    \multicolumn{3}{l}{\it \small\textbf{Zero-shot LMMs} } \\
     LLaVA-OV-0.5B~\cite{li2024llava}                             & 65.1 & 77.4 & \textcolor{gray}{60.0} & 57.1 & 48.5 & 52.8 & 8.01 & 0.9B~/~-\\
     LLaVA-OV-7B~\cite{li2024llava}                               & 70.6 & 85.6 & \textcolor{gray}{68.8} & 70.6 & 52.8 & 60.4 & 98.53 & 8.2B~/~-\\
     \midrule
     \multicolumn{3}{l}{\it \small\textbf{Task-specific models} } \\     
     LLaVA-OV-0.5B-FT                                             & 117.6  & 86.4 & \textcolor{gray}{69.6} & 76.3 & 62.8 &  67.6  & 8.01 & 0.9B~/~35.2M \\
     LLaVA-OV-7B-FT                                               & 124.9  & 90.8 & \textcolor{gray}{75.4} & 89.3 & 72.3 &  77.4  & 98.53 & 8.2B~/~161.5M \\
     \midrule
     
     PAVE-0.5B (w/ audio)                                          & 134.5 &  90.4  & \textcolor{gray}{77.3} & 89.8  & 74.1 & 78.8 & 8.08 & 0.9B~/~41.4M   \\
     PAVE-7B   (w/ audio)                                          & 152.9 & 93.8 & \textcolor{gray}{79.7} & 93.0 & 78.0 &  82.3 & 98.63 & 8.2B~/~170.5M \\ \bottomrule

\end{tabular}
}

\caption{Additional result of PAVE on the audio-visual understanding tasks with audio as additional information. 
\vspace{-1mm}
}  
\label{tab:video_audio_understanding_appendix}  
\end{table*}  

\begin{table*}[t]  
\centering  
\scalebox{0.8}{
\begin{tabular}{lccccc|c|c|c}  
\toprule
\multirow{2}{*}{Method}                                          & \multicolumn{5}{c|}{ScanQA~\cite{azuma_2022_CVPR}} & SQA3D\cite{ma2022sqa3d} & \multirow{2}{*}{TFLOPs} & \multirow{2}{*}{Total / Trainable Params}\\
                                                                 & C     & B-4 & M     & R    & EM@1 & EM@1     &&            \\ 
    \midrule
    \multicolumn{6}{l}{\it \small\textbf{Zero-shot LMMs}} \\
     LLaVA-OV-0.5B~\cite{li2024llava}                            & 17.2  & 1.2 & 13.7  & 18.4 & 0.2 \textcolor{gray}{(28.0)} & 0.8 \textcolor{gray}{(43.0)} & 8.01 & 0.9B~/- \\
     LLaVA-OV-7B~\cite{li2024llava}                              & 91.0  & 5.3 &  18.2 & 45.9 & 26.7 \textcolor{gray}{(44.3)} & 8.3 \textcolor{gray}{(50.7)} &98.53 & 8.2B / - \\
    \midrule
    \multicolumn{6}{l}{\it \small\textbf{Task-specific models}} \\
      LLaVA-OV-0.5B-FT  & 70.5 & 6.5 & 14.3 & 36.9 & 20.5 \textcolor{gray}{(36.3)}  & 44.1 \textcolor{gray}{(45.7)} & 8.01 & 0.9B~/~35.2M\\
      LLaVA-OV-7B-FT    & 95.1 & 13.5 & 19.1 & 47.4 & 27.4 \textcolor{gray}{(46.3)} & 55.8 \textcolor{gray}{(58.1)} & 98.53 & 8.2B / 161.5M \\
    \midrule
     PAVE-0.5B (w/ 3D info)                                       & 84.2 & 13.1 & 17.0 & 42.1 & 23.1 \textcolor{gray}{(40.0)}&  51.1 \textcolor{gray}{(52.8)} & 8.13 & 0.9B~/~41.4M\\
     PAVE-7B   (w/ 3D info)                                       & 103.4 & 16.0 & 19.9 & 49.0 & 29.1 \textcolor{gray}{(48.5)} & 59.0 \textcolor{gray}{(61.4)} & 98.68 & 8.2B / 170.5M\\ \bottomrule

\end{tabular}
}
\caption{Additional result of PAVE on the 3DQA tasks with 3D information as additional information.
}  
\vspace{-1mm}
\label{tab:3d_qa_understanding_appendix}  
\end{table*}

\begin{table*}[t]  
\centering  
\scalebox{0.72}{
\begin{tabular}{l|ccccc|c|c}  
\toprule
     Method                               & ActivtityNet-QA & EgoSchema & NextQA & PerceptionTest & VideoMME (w-subs) & FLOPs (TB) & Total / Trainable Params \\ 
    \midrule
     LLaVA-OV-0.5B~\cite{li2024llava}     & 50.5            & 26.8      & 57.2   & 49.2            & 43.5 & 8.01 & 0.9B~/~- \\
     LLaVA-OV-7B~\cite{li2024llava}       & 56.6            & 60.1      & 79.4   & 57.1            & 61.5 & 98.53 & 8.2B~/~- \\
     \midrule
     PAVE-0.5B (w/ video feature)         & 50.6 & 27.1  &    56.1 & 48.8 & 48.6 & 8.08  & 0.9B~/~41.4M  \\
     PAVE-7B   (w/ video feature)         & 57.1 & 57.4  &    79.6 & 56.0 & 62.9 & 98.63 & 8.2B~/~170.5M \\ \bottomrule

\end{tabular}
}
\caption{Result of PAVE on the additional benchmarks in enhanced video understanding setting. PAVE uses densely sampled video frames as additional information. 
\vspace{-1mm}
}  
\label{tab:general_video_understanding_appendix}  
\end{table*}

\section{Additional Experiment Results} \label{experiment}
\subsection{Results with small Video LLM} 

We now present additional experiment results of PAVE with LLaVA-OneVision 0.5B models for audio-visual QA and 3D QA. 
Table~\ref{tab:video_audio_understanding_appendix} and Table~\ref{tab:3d_qa_understanding_appendix} show the results. PAVE consistently improves the 0.5B and 7B Video LLM's performance by a large margin across both settings. 
This indicates that PAVE effectively leverages additional information when adapting pre-trained Video LLMs into new settings. 

\subsection{Results on Enhanced Video Understanding} 

We present PAVE's result on additional benchmarks in the enhanced video understanding setting. 
Table~\ref{tab:general_video_understanding_appendix} shows PAVE's results in the enhanced video understanding setting with additional benchmarks. PAVE demonstrates a substantial performance gain on VideoMME (w-subtitles). However, we observe only marginal or no improvement on ActivityNet-QA, EgoSchema, NextQA, and PerceptionTest. We hypothesize that this discrepancy may be due to: (1) domain shift—our training data primarily consists of third-person view videos, which may lead to a performance drop in EgoSchema, and (2) the nature of the benchmark questions, which may not require densely temporal information for reasoning.

\begin{table}[t]  
\centering
\resizebox{0.98\linewidth}{!}{
\begin{tabular}{lccc}  
\toprule
       Model                    & Acc. & FLOPs (TB) &  Total / Trainable Params \\
     \midrule
     \multicolumn{4}{l}{\it \small\textbf{Zero-shot LMMs}} \\
     LLaVA-OV-0.5B    &  23.6   & 8.01  & 0.9B~/ -   \\
     LLaVA-OV-7B   &  23.6    & 98.53  & -    \\
     \midrule
     \multicolumn{4}{l}{\it \small\textbf{Task-specific models}} \\
     LLaVA-OV-0.5B-FT             & 28.2     & 8.01  & 0.9B~/~35.2M  \\
     LLaVA-OV-7B-FT             & 29.8     & 98.53  & 8.2B / 161.5M  \\
     TimeSFormer (Ego+Exo)* ~\cite{grauman2024ego}      & 43.7    & -  &   -  \\
     \midrule
     PAVE-0.5B                & 32.4           & 8.15 & 0.9B~/~41.4M \\ 
     PAVE-7B                  & 44.2  & 98.70 & 8.2B / 170.5M \\ 
     \bottomrule

\end{tabular}
}
\vspace{-2mm}
\caption{Performance of PAVE on multi-view video understanding with Ego-Exo4D Demonstrator Proficiency benchmark. LLaVA-OV-7B-FT refers to directly fine-tuning the LLaVA-OneVision on the training set. Our model achieves state-of-the-art performance by only adding a small amount of parameters and FLOPs. * means this baseline may use more training data than PAVE because some of the videos are unavailable to us.} 
\label{tab:ego_exo_videos_appendix}  
\end{table}

\subsection{Results on Multi-view Video Understanding} \label{section_res_multi_view_video}
\noindent\textbf{Motivation and task set up.}
Understanding human activity from video is crucial in many real-world applications, such as augmented reality and robotic learning. Based on the perspective, videos can be broadly classified into ego-centric and exo-centric views. Ego-centric videos capture first-person interactions, focusing on close-up hand-object interactions, while exo-centric videos provide a third-person perspective, recording full-body postures and the surrounding environment. Both perspectives are essential for comprehensive human action understanding. 
Different from the audio-visual QA and 3D QA, where the side-channel information comes from other modalities, in this context, PAVE regards exo-centric videos as side-channel information and integrates it with ego-centric video to adapt the Video LLMs for multi-view video understanding.

\medskip
\noindent\textbf{Training data.} 
We use the training set from the Ego-Exo4D demonstrator proficiency estimation benchmark~\cite{grauman2024ego} as our training data, which consists of 1,904 question-answer pairs. Each pair is associated with one ego-centric video and four exo-centric videos. The task requires the model to classify human action proficiency into one of four categories: Novice, Early Expert, Intermediate Expert, or Late Expert, based on both ego- and exo-centric videos. However, only 1,656 question-answer pairs include the corresponding videos, as the videos for the remaining pairs could not be downloaded due to privacy issues.

\medskip
\noindent\textbf{Implementation details.}
Considering the exo- and ego-centric videos are synchronized along the temporal axis, we sample 32 frames for each of the exo-centric videos. To keep the encoding procedure consistent between the ego- and exo-video, we use the same preprocessing of the LLaVA-OneVision to reshape and crop the video frames. We use SigLIP~\cite{zhai2023sigmoidlosslanguageimage} as the visual encoder and it encodes and downsamples each frame into 196 tokens. We pre-extract the exo-video feature tokens offline to accelerate the training. 
We build PAVE on top of LLaVA-OneVision~\cite{li2024llava} and train the model for 2 epochs.

\medskip
\noindent\textbf{Evaluation benchmark.} 
We use the validation set of the Ego-Exo4D~\cite{grauman2024ego} demonstrator proficiency estimation benchmark for evaluation and report accuracy as the metric. It contains 466 questions and each of the questions is paired with 1 ego-centric video and 4 exo-centric videos.

\medskip
\noindent\textbf{Baselines.}
We use the TimeSFormer (Ego+Exo) from Ego-Exo4D~\cite{grauman2024ego} as our baseline. We also include a baseline that directly fine-tunes the LLaVA-OneVision with LoRA on the training set without using the exo-centric videos, denoted as LLaVA-OV-7B-FT. This baseline allows us to assess whether PAVE can effectively utilize supplementary information.

\medskip
\noindent\textbf{Results.} Table~\ref{tab:ego_exo_videos_appendix} shows the results of PAVE. Compared with the LLaVA-OV-7B-FT, PAVE-7B achieves about 14.4\% improvement by adding only 9M parameters and 0.17 TFLOPs during inference. This big improvement indicates that the exo-centric videos provide crucial additional information for human action understanding.
Moreover, PAVE achieves state-of-the-art performance on the demonstrator proficiency estimation benchmark, substantiating that  PAVE can adapt a pre-trained Video LLM to an unseen setting by leveraging supplementary information. 



\section{Implementation and Experiment Details} \label{implementation}
We first describe the general implementation detail of the PAVE in Section~\ref{pave_inple}. Then, we describe the experiment details for 3 settings considered in the main paper, including audio-visual QA (Section~\ref{avqa}), 3DQA (Section~\ref{3dqa}), and enhancing video QA (Section~\ref{enhanced_video_qa}). We also demonstrate how we calculate the Flops for the model in Section~\ref{flops_calc}.

\subsection{Implementation Detail of PAVE} \label{pave_inple}

Inside the temporal-aligned cross-attention layer, we add rotary position embedding to the query and key tokens. Specifically, we apply different rotary positional embedding according to the layout of side-channel tokens $\mathbf{z}^s$. We mainly consider two types of $\mathbf{z}^s$: (a) $\{\mathbf{z}^s\}$ includes both spatial and temporal dimensions, such as tokens from video backbones or from a 3D backbone; and (b) $\{\mathbf{z}^s\}$ only contains temporal dimension, such as audio tokens. For the first case, we will add 3D rotary positional embedding (along the temporal, height, and width dimensions). For the second case, we will only add rotary positional embedding along the temporal axis. After cross-attention, we use a two-layer MLP, followed by a layer norm. 
After the PAVE layers, we add another two-layer MLP, followed by a layer norm, as the adapter. We initialize the $\gamma$ in the layer norm to zero. 


\subsection{Audio-Visual QA}\label{avqa}
In this setting, the input of the PAVE has two parts: (1) the visual tokens $\mathbf{z}^v$ from the Video LLM's visual encoder, and (2) the audio tokens $\mathbf{z}^s$ from a side-channel signal encoder.

\medskip
\noindent \textbf{Visual Encoder.} For $\mathbf{z}^v$, we follow the default setting used in LLaVA-OneVision~\cite{li2024llava}. We uniformly sample 32 frames from the video and use the same preprocessing of the LLaVA-OneVision to reshape and crop the video frames. We use SigLIP~\cite{zhai2023sigmoidlosslanguageimage} as the visual encoder and it encodes and downsamples each frame into 196 tokens.

\medskip
\noindent \textbf{Side-Channel Signal Encoder.}
For $\mathbf{z}^s$, we follow the pre-processing step of ImageBind~\cite{girdhar2023imagebind}, which resamples the audio at 16KHz. 
We segment the audio into overlapping 2-second clips with a 1-second stride and encode each clip using the audio encoder of ImageBind. This process generates a 1024-dimensional audio token for every 1 second of the audio signal.
Since we do not fine-tune the audio encoder, we extract the audio feature tokens offline in order to accelerate the training.

\medskip
\noindent \textbf{Network Architecture.} For the PAVE design, we use 2 cross-attention layers with hidden dimension 512 and have 4 attention heads. 
For LoRA layers in the LLM, we use LoRA$\_r$ = 64 and LoRA$\_\alpha$ = 16. 

\medskip
\noindent \textbf{Training Details.}
For training, we use AdamW~\cite{loshchilov2019decoupledweightdecayregularization} optimizer with a linear warmup using the first 3\% of iterations. We use the cosine annealing learning rate during the training. We set the base learning rate as 2e-5 and the batch size as 32. All the experiments are run on 2 A100 80G GPUs. 

\medskip
\noindent\textbf{Training data.} We choose the open-end QA dataset AVSD~\cite{alamri2019audiovisualsceneawaredialog}, and closed-end QA dataset AVQA~\cite{yang2022avqa} and Music-AVQA~\cite{Li2022Learning} as training dataset. AVSD contains 79k question-answer pairs across 7,985 videos with each paired 10 questions. AVQA has 40k question-answer pairs coupled with 40k Videos. Music-AVQA consists of 32k question-answer pairs and 9277 videos.

\medskip
\noindent\textbf{Evaluation benchmark.} We follow the protocol in previous works~\cite{ye2024catenhancingmultimodallarge, pham2022videodialogconversationobjects} to evaluate PAVE. For AVSD, we use the AVSD@DSTC7 test split and report CIDEr score as the metric. This benchmark consists of 1,000 audio-visual questions.
We use COCO API~\cite{lin2015microsoftcococommonobjects} to calculate the CIDEr score between the model predictions and the ground truth answers. 
For AVQA, we evaluate PAVE on the eval split and report the accuracy as the metric. This benchmark contains 17k questions that require reasoning based on audio and visual information.
For Music-AVQA, we evaluate PAVE on the test split and report the accuracy as the metric. This benchmark contains 9185 questions, which can be categorized into visual, audio, and audio-visual questions.


\subsection{3DQA} \label{3dqa}
In this setting, the input of the PAVE consists of two parts: (1) the visual tokens $\mathbf{z}^v$ from the Video LLM's visual encoder, and (2) the 3D tokens $\mathbf{z}^s$ from a side-channel signal encoder.

\medskip
\noindent \textbf{Visual Encoder.} For $\mathbf{z}^v$, we use the same setting as the one in Section~\ref{avqa}. 

\medskip
\noindent \textbf{Side-Channel Signal Encoder.} For encoding the side-channels information into $\mathbf{z}^s$, we utilize the 3D encoder which contains two parts 1. a visual encoder which encodes the RGB frames into visual feature tokens. 2. a spatial embedding that adds the encoded 3D information on the visual feature tokens. We uniformly extract 32 RGB-D frames from the scan and use ViT~\cite{dosovitskiy2021imageworth16x16words} to extract the visual features from the RGB frames. We then add spatial embeddings to visual features following the LLaVA-3D~\cite{zhu2024llava3dsimpleeffectivepathway} by making use of the depth information and the camera pose. It generates 576 tokens for each frame, with a token dimension of 1024. We pre-extract the 3D feature to accelerate the training.

\medskip
\noindent \textbf{Network Architecture and Training Details.} For the PAVE design and the training configuration, we use the same hyper-parameters used in Section~\ref{avqa}.


\medskip
\noindent\textbf{Training data.} For 3D QA tasks, we consider ScanQA~\cite{azuma_2022_CVPR} and SQA3D~\cite{ma2022sqa3d}. ScanQA and SQA3D contain 25K and 26K training question-answer pairs, respectively. They share the same scanning data set which contains 562 3D scanning from ScanNet~\cite{dai2017scannet}.

\medskip
\noindent\textbf{Evaluation benchmark.} We report our model performance on the ScanQA validation set, which contains 4,675 questions covering both object position reasoning and object recognition, and the SQA3D test set with 3519 questions, which consists of 5 different types of questions.
Following previous work~\cite{zhu2024llava3dsimpleeffectivepathway}, we report the CIDEr (C), BLEU-4 (B-4), METEOR (M), ROUGE(R), and top-1 Exact Match (EM@1) metrics on ScanQA and report EM@1 on SQA3D. We use the evaluation pipeline set up by LLaVA-3D to evaluate our model on ScanQA and SQA3D. 







\subsection{Enhancing Video QA} \label{enhanced_video_qa}
In this setting, the input of the PAVE has two parts: (1) the visual tokens from the Video LLM's visual encoder $\mathbf{z}^v$, extracted at sparsely sample video key frames, and (2) the side-channel visual tokens $\mathbf{z}^s$, derived from a high frame rate video.

\medskip
\noindent \textbf{Visual Encoder.} For $\mathbf{z}^v$, we use the same setting as the one in Section~\ref{avqa}.  

\medskip
\noindent \textbf{Side-Channel Signal Encoder.} In this case, the side-channel signals $\mathbf{z}^s$ are high frame rate videos. We sample the video frames at the frame rate of 2fps and use the default pre-processing step of the LanguageBind to reshape and crop the video frames. To leverage LanguageBind~\cite{zhu2023languagebind} to encode the high-frame-rate video frames, we split the video frames along the temporal axis into multiple non-overlap groups with each group containing 8 frames. We later concatenate the encoded features of all groups along the temporal axis. To reduce the overhead of the PAVE, inspired by the Slow-Fast~\cite{feichtenhofer2019slowfastnetworksvideorecognition}, we downsample the spatial resolution of the video feature of each video frame from 16 $\times$ 16 to 2 $\times$ 2. We do not utilize the classification tokens from the output of the LanguageBind. Since we do not fine-tune LanguageBind's video encoder, we pre-extract the video features in order to speed up the training.

\medskip
\noindent \textbf{Network Architecture and Training Details.} For the PAVE design and the training configuration, we use the same hyper-parameters used in Section~\ref{avqa}.



\medskip
\noindent\textbf{Training data.} We create a subset from LLaVA-Video-178K \cite{zhang2024videoinstructiontuningsynthetic} by first sampling all videos longer than 1 minute and then randomly choosing 2 question-answer pairs for each video. This process creates a training set that contains 57K videos and 114K question-and-answer pairs.

\medskip
\noindent\textbf{Evaluation benchmark.} We use VideoMME~\cite{fu2024video}, MVBench~\cite{li2023mvbench}, and MLVU~\cite{zhou2025mlvubenchmarkingmultitasklong} as evaluation benchmarks. VideoMME and MVBench are both comprehensive video benchmarks and cover different types of subtasks, while MLVU focuses on long video understanding.
VideoMME includes 6 key domains and 30 sub-classes. It contains 900 videos, ranging from less than one minute to nearly one hour. There are 2,700 questions with each accompanied by four options. 
MVBench includes 20 different sub-tasks, such as object shuffling and fine-grained pose estimation, which require detailed temporal information. In total, it has about 4000 questions and 3900 videos. 
MLVU contains 2175 questions and 1337 long videos.
All benchmarks adopt accuracy as the performance metric.

\subsection{The Calculation of Inference FLOPs.} \label{flops_calc}
We now describe how the floating-point operations (FLOPs) are reported in our experiments.
Since the visual-encoder and the side-channel information encoder are replaceable modules in PAVE settings (i.e. we can use encoder with different scales at different settings.), we only consider the FLOPs of PAVE and LLM, provided by the LLM-Viewer~\cite{yuan2024llm}. During the FLOPs calculation of LLM, we consider 6272 visual tokens, and following the previous work~\cite{shang2024llavaprumergeadaptivetokenreduction}, we add 40 additional tokens for the text. We then calculate and add the FLOPs of PAVE. 

%

The FLOPs of PAVE is calculated as follows. The input of the PAVE consists of two parts, the visual tokens $\mathbf{z}^v$ from the Video LLM's visual backbone, and the side-channel information tokens $\mathbf{z}^s$. We consider the case that the visual tokens $\mathbf{z}^v$ come from 32 video frames and Video LLM's visual backbone generates 196 tokens for each frame. 
\begin{itemize}
    \item \textbf{Audio-visual QA}: We assume the length of the video at inference time is 2 minutes and the audio encoder will generate 1 token for each second of the audio. It yields 120 audio tokens. The cross-attention is conducted over 196 query tokens and 4 key tokens. PAVE thus introduces about 0.07 TB and 0.10 TB FLOPs for 0.5B and 7B models, respectively.  
    \item \textbf{3D QA}: We uniformly sample 32 frames and send them into the 3D backbone. It generates 576 tokens for each frame and yields 18432 tokens in total. The cross-attention is conducted over 196 query tokens and 576 key tokens. PAVE introduces about 0.12 TB and 0.15 TB FLOPs for 0.5B and 7B models, respectively.
    \item \textbf{Enhancing video QA}: We assume the length of the video at inference time is 2 minutes---close to the average duration of videos on VideoMME and MVBench. We sample the frames at 2 fps and sent them to the video backbone. We down-sample the tokens of each frame spatially to 2 by 2 grids. It produces 960 video tokens in total. The cross-attention is conducted over 196 query tokens and 30 key tokens. PAVE adds about 0.07 TB and 0.10 TB FLOPs for 0.5B and 7B models, respectively.
    \item \textbf{Multi-view Video Understanding}: We uniformly sample 32 frames for each exo-centric video and send them into the SigLIP. It generates 196 tokens for each frame and yields 25,088 tokens for 4 exo-centric videos in total. The cross-attention is conducted over 196 query tokens and 784 key tokens. PAVE adds about 0.14 TB and 0.17 TB FLOPs for 0.5B and 7B models, respectively.  
\end{itemize}

\section{Additional Visualization} \label{visualization}

\begin{figure*}[t!]
    \centering
    \includegraphics[width=0.75\linewidth]{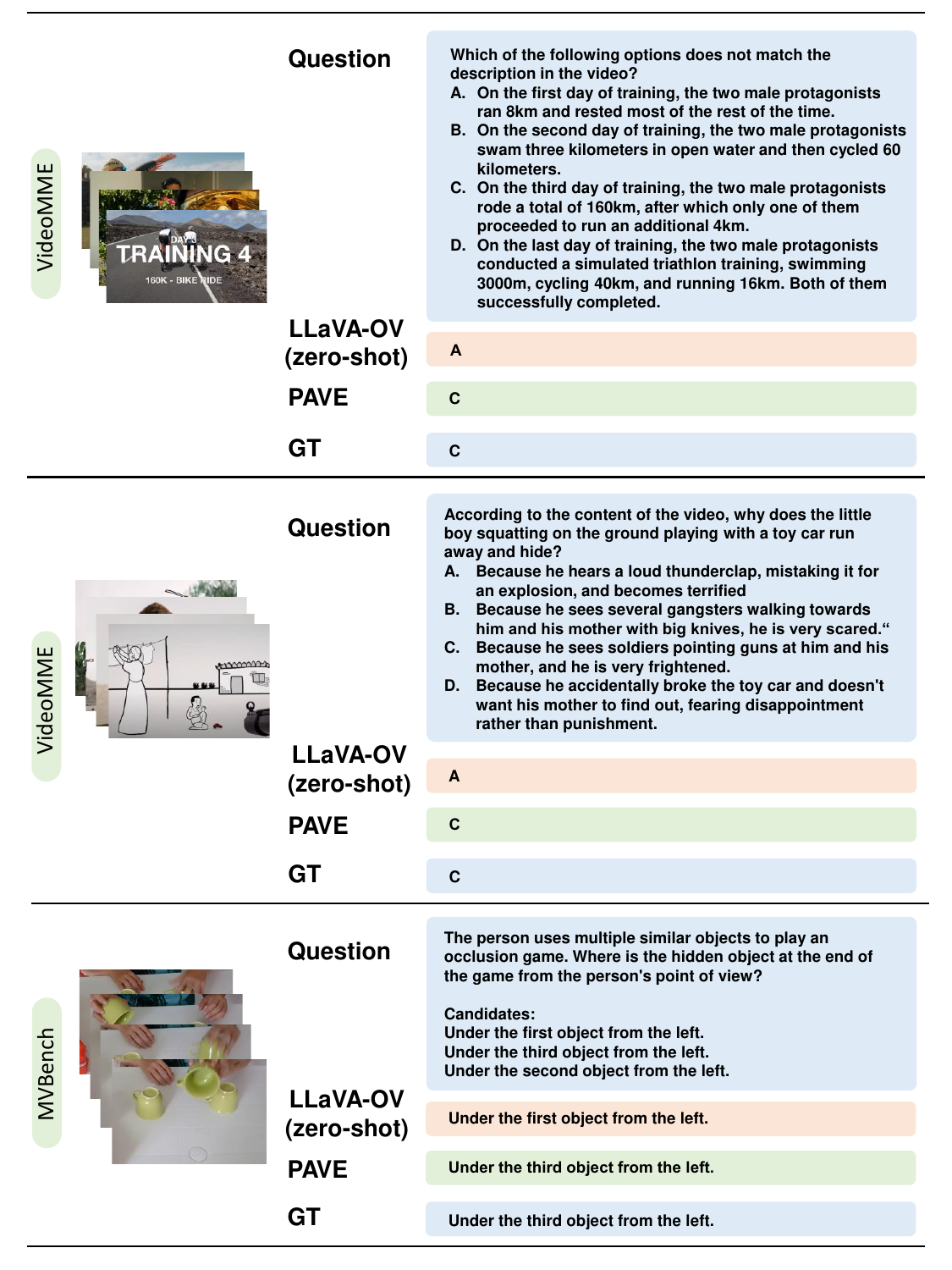}
    \vspace{-1em}
    \caption{Visualization of the QA results on enhanced video QA task. By making use of the video feature of the densely sampled video frames, PAVE captures more details in the video and thus improves the performance of video understanding.}
    \label{fig:enhanced_qa_visualization}
    \vspace{-1.5em}
\end{figure*}

We present additional visualization of the PAVE's results for enhanced video QA in Figure~\ref{fig:enhanced_qa_visualization} with videos from VideoMME~\cite{fu2024video} and MVBench~\cite{li2023mvbench}.





\end{document}